%% file: acl_latex.tex
\documentclass[11pt]{article}
\PassOptionsToPackage{table}{xcolor}
\usepackage{booktabs} 
\usepackage{longtable}
\usepackage{pifont}
\usepackage[final]{acl}
\usepackage{times}
\usepackage{latexsym}
\usepackage[T1]{fontenc}
\usepackage[utf8]{inputenc}
\usepackage{microtype}
\usepackage{inconsolata}
\usepackage{graphicx}
\usepackage{multirow}
\usepackage{multicol}
\usepackage{pgf}
\usepackage{setspace}
\usepackage{stfloats}
\usepackage{subcaption}
\usepackage{svg}
\usepackage{tabularx}
\usepackage{tcolorbox}
\usepackage{times}
\usepackage{url}
\usepackage{color-edits}
\usepackage{amsmath}
\usepackage{marvosym}
\usepackage{enumitem}
\usepackage{amsfonts}

\usepackage[normalem]{ulem}
\useunder{\uline}{\ul}{}

\definecolor{cellbase}{RGB}{245, 250, 245}    
\definecolor{cellmid}{RGB}{220, 240, 220}     
\definecolor{cellgood}{RGB}{190, 225, 190}    
\definecolor{cellbest}{RGB}{150, 210, 150}

\newcommand{\cellcolorbyimprove}[2]{
  \pgfmathsetmacro{\improve}{(#2-#1)/max(1,#1)*100}
  \ifdim \improve pt > 8pt
    \cellcolor{cellbest} #2
  \else\ifdim \improve pt > 3pt
    \cellcolor{cellgood} #2
  \else\ifdim \improve pt > 1pt
    \cellcolor{cellmid} #2
  \else
    \cellcolor{cellbase} #2
  \fi\fi\fi
}

\newcommand{\cellcolorbyimprovenew}[2]{
  \pgfmathsetmacro{\improve}{(#2-#1)/max(1,#1)*100}
  \ifdim \improve pt > 20pt
    \cellcolor{cellbest} #2
  \else\ifdim \improve pt > 10pt
    \cellcolor{cellgood} #2
  \else\ifdim \improve pt > 5pt
    \cellcolor{cellmid} #2
  \else
    \cellcolor{cellbase} #2
  \fi\fi\fi
}
\makeatletter
\def\@fnsymbol#1{}
\makeatother

\title{DARL: Encouraging Diverse Answers for General Reasoning \\without Verifiers}

\author{
Chongxuan Huang\textsuperscript{\rm{1,*}}\thanks{\textsuperscript{*}\ Work done during an internship at Kuaishou Technology.}, 
Lei Lin\textsuperscript{\rm{2,$\dagger$,$\ddagger$}}\thanks{\textsuperscript{$\dagger$}\ Project Lead.}, 
Xiaodong Shi\textsuperscript{\rm{1,3,$\ddagger$}}\thanks{\textsuperscript{$\ddagger$}\ Corresponding author.},
Wenping Hu\textsuperscript{\rm{2}}, Ruiming Tang\textsuperscript{\rm{2}}\\
\textsuperscript{1} School of Informatics, Xiamen University \\
\textsuperscript{2} Kuaishou Technology, Beijing, China \\
\textsuperscript{3} Key Laboratory of Digital Protection and Intelligent Processing of Intangible Cultural \\Heritage of Fujian and Taiwan (Xiamen University), Ministry of Culture and Tourism \\
\texttt{huangchongxuan@stu.xmu.edu.cn,mandel@xmu.edu.cn} \\
\texttt{\{linlei,huwenping,tangruiming\}@kuaishou.com}
}

\begin{document}
\maketitle
\begin{abstract}
\input{chapter/1_abstract}

\end{abstract}

\input{chapter/2_introduction}

\input{chapter/3_RelatedWork}

\input{chapter/4_Background}

\input{chapter/5_Method}

\input{chapter/6_ExperimentalSetup}

\input{chapter/7_ResultsandAnalysis}

\input{chapter/8_Conclusion}

\input{chapter/9_Limitations}



\bibliography{custom}

\clearpage
\appendix

\input{chapter/11_appendix}


\end{document}

%% file: chapter/1_abstract.tex
Reinforcement Learning with Verifiable Rewards (RLVR) has demonstrated promising gains in enhancing the reasoning capabilities of large language models. However, its dependence on domain-specific verifiers significantly restricts its applicability to open and general domains. Recent efforts such as RLPR have extended RLVR to general domains, enabling training on broader datasets and achieving improvements over RLVR. However, a notable limitation of these methods is their tendency to overfit to reference answers, which constrains the model’s ability to generate diverse outputs. This limitation is particularly pronounced in open-ended tasks such as writing, where multiple plausible answers exist. To address this, we propose \textbf{\emph{DARL}}, a simple yet effective reinforcement learning framework that encourages the generation of diverse answers within a controlled deviation range from the reference while preserving alignment with it. Our framework is fully compatible with existing general reinforcement learning methods and can be seamlessly integrated without additional verifiers. Extensive experiments on thirteen benchmarks demonstrate consistent improvements in reasoning performance. Notably, DARL surpasses RLPR, achieving average gains of 1.3 points on six reasoning benchmarks and 9.5 points on seven general benchmarks, highlighting its effectiveness in improving both reasoning accuracy and output diversity.

%% file: chapter/2_introduction.tex
\section{Introduction}

Recent advances in Large Reasoning Models (LRMs), such as DeepSeek-R1 \cite{guo2025deepseek} and OpenAI's O1 \cite{jaech2024openai}, have demonstrated that reasoning capabilities are crucial for improving model performance across various tasks \cite{hu2025open,liu2025understanding,luo2025deepcoder}. To this end, Large-Scale Reinforcement Learning with Verifiable Rewards (RLVR) has emerged as an effective training paradigm for LRMs~\cite{xie2025logic}.
Specifically, RLVR assigns rewards solely based on final-answer correctness, encouraging the emergence of intermediate reasoning without explicit supervision. By leveraging clear and verifiable reward signals, RLVR guides models to discover reasoning trajectories that reliably lead to correct solutions.

However, RLVR exhibits notable limitations when compared to LLMs pretrained on general data, which acquire broad foundational abilities \cite{zhang2025survey}. While RLVR performs well in domains with well-defined, objectively correct answers, such as mathematics \cite{hu2025open,liu2025understanding,zeng2025simplerl} and programming \cite{luo2025deepcoder}, it struggles in areas requiring subjective judgment, including creative writing or open-ended question answering. This limitation arises because RLVR relies on manually designed verifiers, which are difficult to construct for complex, subjective tasks . The need for domain-specific experts to create these verification systems also results in high costs and resource requirements \cite{yu2025rlpr}. Moreover, the binary reward structure inherent to RLVR, which provides sparse feedback, limits its ability to handle more nuanced or general tasks \cite{zhang2025survey}.

To address these limitations, recent research has explored several directions for extending RLVR to general reasoning tasks. A common strategy leverages generative models to produce fine-grained feedback \cite{mahan2024generative,li2023generative,zheng2023judging}, mitigating the limitations of rule-based verifiers, yet it demands large-scale annotations and complex verifier models, which increase computational cost and risk reward hacking, as observed in DeepSeek-V3 and DeepSeek-R1 \cite{guo2025deepseek}. To reduce dependence on labeled data, unsupervised reward modeling extracts signals directly from model-internal consistency, entropy, or self-generated knowledge (RLIF \cite{zhao2025learning}, EM-RL \cite{agarwal2025unreasonable}, EMPO \cite{zhang2025right}, CoVo \cite{zhang2025consistent}, RENT \cite{prabhudesai2025maximizing}, TTRL \cite{zuo2025ttrl}, SeRL \cite{fang2025serl}). Although scalable, such closed-loop training often suffers from instability and excessive reliance on pretrained priors. More recent hybrid methods, such as NOVER \cite{liu2025nover}, VeriFree \cite{zhou2025reinforcing}, RLPR \cite{yu2025rlpr}, JEPO \cite{tang2025beyond}, and LaTRO \cite{chen2024language}, combine self-generated signals with reference answers to build more accurate reward functions. These designs yield more reliable rewards and have achieved leading performance across several reasoning benchmarks. However, generating diverse yet valid outputs is critical for capturing underlying semantic and reasoning structures and for improving generalization, and existing methods tend to overfit to the reference answers, thereby restricting the model’s capacity to produce such diversity.

Motivated by these observations, we argue that an effective general reinforcement learning framework for reasoning should avoid overfitting to specific reference answers or relying on explicit verifiers. Instead, it should encourage controlled exploration beyond the reference, allowing the model to generate diverse yet reference-consistent answers. To this end, we propose \textbf{\emph{DARL}}, a verifier-free, simple yet effective reinforcement learning framework that promotes answer diversity within a bounded deviation range from the reference answer, without sacrificing reference consistency. Our main contributions are summarized as follows:

\begin{itemize}

\item First, we propose a general verifier-free reinforcement learning framework that encourages models to generate diverse answers while remaining consistent with the reference. Unlike existing reference-based methods that tend to overfit exact ground-truth expressions, our approach explicitly promotes controlled exploration over alternative answer formulations without compromising alignment with the reference.

\item  Second, we show that DARL significantly improves answer diversity. At the policy level, our method maintains higher policy entropy during training, indicating enhanced and more stable exploration. At the output level, likelihood-based analysis shows that DARL assigns higher probability mass to semantically equivalent but lexically diverse answers compared to RLPR. These results demonstrate that our approach promotes controlled answer diversity, which is especially beneficial for open-ended tasks such as writing.

\item Third, DARL consistently improves performance across diverse reasoning and general-domain benchmarks. It achieves notable gains on benchmarks with multiple valid reasoning paths (e.g., GPQA and AIME24), while maintaining comparable performance on more constrained tasks such as MMLU-Pro. Moreover, DARL delivers substantial improvements in general-domain settings, with average gains of 11.57 points on logical reasoning, 12.44 points on code generation, and 3.35 points on writing tasks, demonstrating strong generalization in open-ended scenarios.
\end{itemize}

%% file: chapter/3_RelatedWork.tex
\section{Related Work}
\paragraph{Language Model Reasoning.}

Reinforcement learning has repeatedly shown that narrow and well-defined reward signals can effectively guide agents to achieve superhuman performance on complex tasks \cite{zhang2025survey}. As the primary training signal, rewards determine the optimization trajectory of the model and shape its behavior. In the context of large-scale language model training, rule-based verifiable rewards have been widely adopted to provide scalable and reliable supervision, particularly in domains such as mathematics \cite{hu2025open,liu2025understanding,zeng2025simplerl} and code generation \cite{luo2025deepcoder}. This approach, formalized as RLVR, replaces learned reward models with programmatic verifiers that emit binary feedback in domains where correctness can be objectively assessed. For example, DeepSeek-V3 \cite{liu2024deepseek} integrated task-specific rule-based reward systems, while DeepSeek-R1 extended this framework by incorporating both accuracy-based and format-based reward mechanisms. Subsequent studies, including Logic-RL \cite{xie2025logic}, DAPO \cite{yu2025dapo}, SimpleRL \cite{zeng2025simplerl}, OpenReasoner \cite{hu2025open}, Dr.GRPO \cite{liu2025understanding}, and VAPO \cite{yue2025vapo}, have investigated best practices for balancing exploration and exploitation, the emergence of “aha” moments, and the influence of task difficulty variations on learning dynamics.

\paragraph{Verifier-Free Reasoning.}

Generative reward modeling and verifier-free approaches have recently emerged as complementary directions for scaling reinforcement learning for LLM reasoning. Early work demonstrated that learned, generation-based verifiers can provide finer-grained, model-aware feedback compared to brittle rule-based checks (e.g., GenRM \cite{mahan2024generative}). However, training standalone reward models typically requires large, complex annotation corpora, increasing system complexity and computational cost \cite{su2025crossing,lu2025writing}.

To reduce reliance on expensive labels, a growing body of research investigates unsupervised or self-supervised reward signals that derive intrinsic feedback from model consistency, confidence, or self-generated knowledge \cite{zhang2025survey}. Representative techniques include entropy-minimization and mutual-information objectives, as well as self-scoring and voting schemes. These methods scale without human annotation and can enhance apparent performance, but they are prone to reward-hacking, overconfidence, entropy collapse, and excessive reliance on pretraining priors, potentially limiting true reasoning generalization \cite{zhang2025right, cui2025entropy, franken2024self, zuo2025ttrl}.

A middle-ground approach leverages ground-truth supervision to enhance self-generated signals, producing more reliable rewards (e.g., JEPO \cite{tang2025beyond}, LaTRO \cite{chen2024language}, NOVER \cite{liu2025nover}, VeriFree \cite{zhou2025reinforcing}, RLPR \cite{yu2025rlpr}). While these methods guide models toward reference-aligned answers, they often induce rapid entropy collapse and overfit to specific solution patterns. Our method extends this paradigm by promoting answer diversity: it encourages the model to generate answers that deviate moderately from the ground truth within a controlled range, thereby balancing reliability and diversity in reward optimization.

%% file: chapter/4_Background.tex
\section{Background}
\paragraph{Rule-based Reward.}
Reinforcement Learning from Verifiable Rewards is a general post-training paradigm that leverages verifiable indicators—such as answer correctness or code execution success—as reward signals. This approach enables models to exhibit capabilities such as planning, reflection, and self-correction in long-chain reasoning tasks \cite{liu2025understanding,xie2025logic}.

Given a prompt that requires the model to generate intermediate reasoning steps followed by a final answer ( y ), the training objective is to optimize the policy model $\pi_{\theta}$ via reinforcement learning, using a verifiable reward $R_{\mathrm{rule}}$.
The reward is computed individually by a rule-based verifier $v: \mathcal{A} \to \{0,1\}$, which evaluates the correctness of the final answer:
\begin{equation}\label{eq:rule_reward}
R_{\mathrm{rule}} = v(y)
\end{equation}
While the outcome reward is inherently sparse, it offers accurate supervision, guiding the model to autonomously produce intermediate reasoning tokens that facilitate the generation of a more reliable final answer.

\paragraph{Verifier-Free Reward.}

While RLVR has shown strong performance on tasks with explicit verifiers, such as mathematical problem solving and code generation, its applicability to open-domain reasoning is limited, as human-designed verifiers are infeasible for diverse and context-dependent natural language. To address this, recent {verifier-free} approaches propose generalizable reward functions, leveraging the assumption that {high-capacity models inherently produce consistent and self-evaluable outputs}, thereby enabling reward estimation without external supervision and enhancing scalability.

Formally, given a reasoning prompt \(x\), the policy model generates a composite output \(o = z \oplus y\), where \(z\) represents the intermediate reasoning process and \(y\) the final answer. The reward \(r\) is computed from the model’s {intrinsic confidence} in the reference answer \(y^{*}\) conditioned on the reasoning chain \(z\), yielding a dense and differentiable training signal:
\begin{equation}\label{eq:verifier_free_reward}
\nabla_{\theta} J_{\mathrm{VerifierFree}}(\theta) = \mathbb{E}_{o \sim \pi_{\theta}(\cdot|x)} \big[ r \cdot \nabla_{\theta} \log \pi_{\theta}(o|x) \big]
\end{equation}
Here, \(r\) is derived from {model-internal signals} rather than external verifiers. For example, RLPR \cite{yu2025rlpr} computes \(r\) based on the {likelihood of generating the ground-truth answer given the intermediate reasoning chain}, effectively aligning reward estimation with the model’s probabilistic reasoning process. This verifier-free paradigm broadens the applicability of RLVR to general reasoning tasks while mitigating the reliance on domain-specific verifiers and human annotations.

%% file: chapter/5_Method.tex
\section{Method}

\begin{figure*}[t!]
    \centering
    \includegraphics[width=1.0\linewidth]{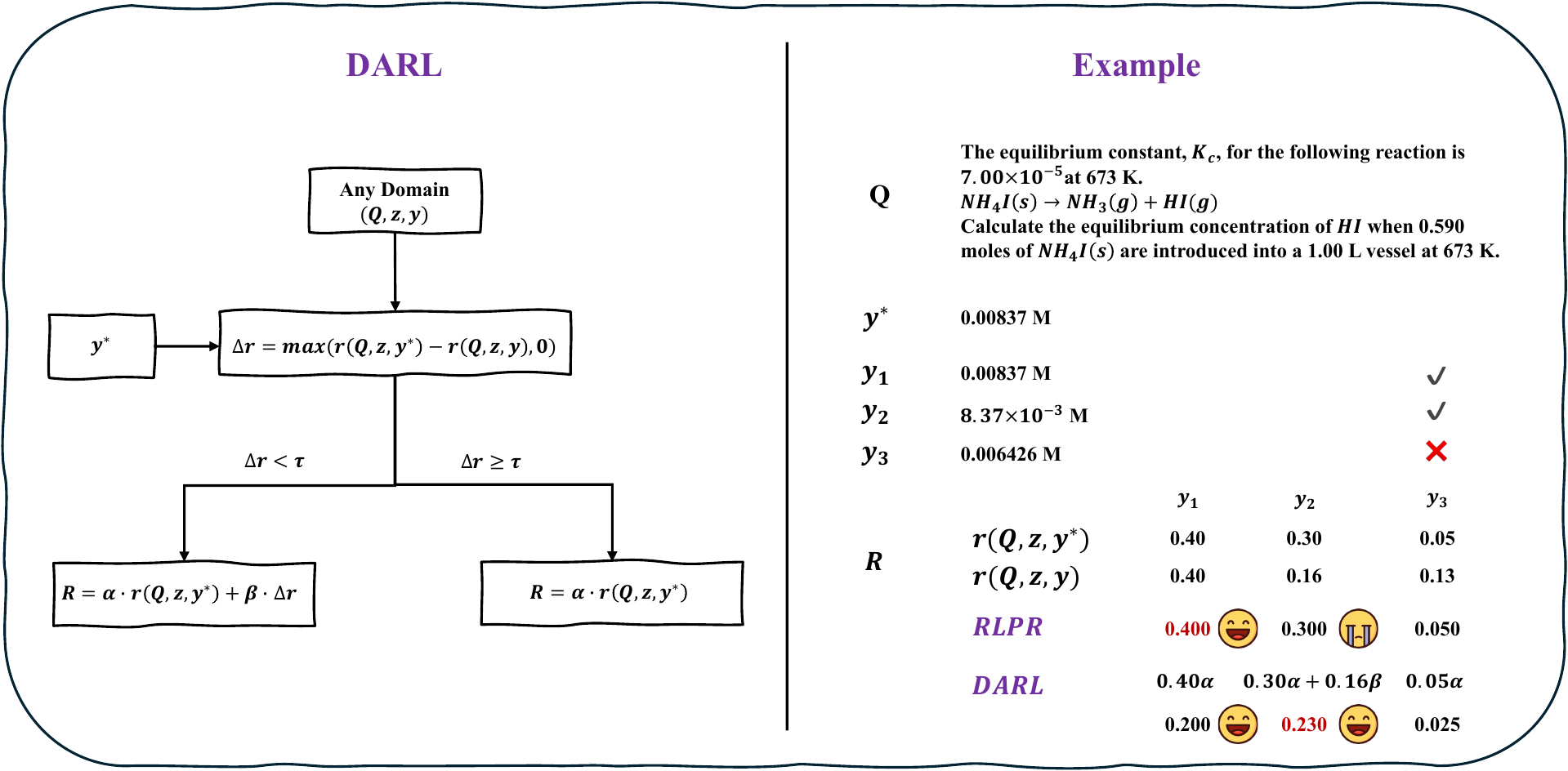}
    \caption{Overview of DARL. $Q$: input question, $z$: generated reasoning content before final answer, $y$: generated final answer, $y^*$: reference answer.}
    \label{fig:paper_total}
\end{figure*}

To promote the generation of diverse answers, we introduce a dynamic, diversity-aware reward that adaptively adjusts the reward threshold for answer diversity according to the model’s confidence.

Given an input $x$, the model generates a reasoning trace $z$ followed by an answer $y$, while $y^{\ast}$ denotes the ground-truth answer. The diversity reward is defined as:
\begin{equation}
\Delta r = \max\big( r(y^\ast, z, x) - r(y, z, x), 0 \big)
\end{equation}

where $r(\cdot)$ is a reward function derived from model-internal signals. Here, $\Delta r$  quantifies the deviation between the model-generated answer and the reference answer. When both $r(y^\ast, z, x)$ and $r(y, z, x)$ are relatively high and their difference falls within a predefined range, it indicates that, given the current reasoning trace $z$, the model has sufficient capability to generate $y^\ast$. Consequently, the answer $y$ produced under the same reasoning trace is also likely to be correct. In other words, when $\Delta r$ falls within the acceptable range, $y$ and $y^\ast$ are highly likely to be consistent answers. Under this condition, the model is encouraged to explore alternative expressions that differ from the reference, and deviations within this range are rewarded, thereby promoting answer diversity while maintaining consistency with the reference answer.

The overall reward combines a reference-aligned term with a bounded answer diversity component:
\begin{equation}
\bar{r} = \alpha \cdot r(y^\ast, z, x) + \beta \cdot \Delta r \cdot \mathbf{1}\Big[ \Delta r \leq \tau \Big]
\end{equation}
where $\alpha$ and $\beta$ are trade-off coefficients controlling the balance between exploitation and answer diversity (with $\alpha+\beta=1$), and $\tau$ denotes a fixed diversity threshold. In our implementation, the reward $r(\cdot)$ is computed as the average token probability, following the approach used in RLPR.
\begin{equation}
r(y,z,x)
=\frac{1}{|y|}\sum_{i=1}^{|y|}
p_{\theta}\!\left(y_i \mid x, z, y_{<i}\right)
\end{equation}
To further enhance adaptability, we introduce a dynamic thresholding strategy that scales the diversity boundary according to the model’s confidence in $r(y^{\ast}, z, x)$:
\begin{equation}
\bar{r} = \alpha \cdot r(y^\ast, z, x) + \beta \cdot \Delta r \cdot \mathbf{1}\Big[ \Delta r \leq \frac{r(y^\ast, z, x)}{\gamma} \Big]
\end{equation}

When the model assigns high confidence to the reference-aligned reward, it indicates sufficient capability to generate answers consistent with the reference. In this regime, we expand the allowable diversity margin to encourage exploration of alternative answers that differ in expression while remaining consistent with the reference. Conversely, when the model’s confidence is low, the diversity range is restricted, biasing learning toward reproducing the reference answer. This prevents premature diversification and stabilizes training when the model’s ability to generate correct answers is limited. Overall, this dynamic answer diversity reward adaptively balances reference alignment and controlled exploration, enabling the generation of multiple reference-consistent solutions while preserving correctness, thereby enhancing both generalization and robustness.

%% file: chapter/6_ExperimentalSetup.tex
\section{Experimental Setup}
\paragraph{Dataset.}

We utilize the dataset collection released with RLPR \citep{yu2025rlpr}, which was originally curated from the WebInstruct corpus by \citet{ma2025general}. The dataset consists of 77.7K training instances of high-quality, non-mathematical reasoning prompts spanning diverse domains, including commonsense reasoning, factual consistency, multi-hop inference, and abstract problem solving.

\paragraph{Models.}
Our experiments primarily employ {Llama3.1} \citep{dubey2024llama} and {Qwen2.5} \citep{Yang2024Qwen25TR} as the backbone policy models. Both models have demonstrated strong zero-shot and instruction-following capabilities across a wide range of reasoning benchmarks. Unless otherwise specified, we report results for Llama3.1-8B-Instruct and Qwen2.5-7B-Base variants.

\paragraph{Baselines.}
For fair and consistent comparison, we adopt the same set of baseline methods as RLPR \citep{yu2025rlpr}, following its official training configurations. Specifically, our comparisons cover (1) base and instruct models (e.g., Qwen2.5 and Llama3.1), (2) reinforcement learning approaches with rule-based, implicit, or verifier-based rewards (e.g., PRIME \cite{cui2025process}, SimpleRL-Zoo \cite{zeng2025simplerl}, Oat-Zero \cite{liu2025understandingr1zeroliketrainingcritical}, and General Reasoner \cite{ma2025general}), and (3) concurrent verifier-free reinforcement learning methods such as TTRL \cite{zuo2025ttrl}, VeriFree \cite{zhou2025reinforcing} and RLPR \cite{yu2025rlpr}. All implementations and hyperparameters are consistent with those reported in RLPR, ensuring a fair comparison.

\paragraph{Evaluation.}
We evaluate all models on a diverse set of benchmarks covering knowledge and factual reasoning, mathematical and symbolic reasoning, logical inference, code generation, and generative writing. The evaluation suite comprises MMLU‑Pro \cite{wang2024mmluprorobustchallengingmultitask}, GPQA \cite{rein2023gpqagraduatelevelgoogleproofqa}, TheoremQA \cite{chen2023theoremqatheoremdrivenquestionanswering}, MATH‑500 \cite{cobbe2021math}, Minerva \cite{lewkowycz2022minerva}, AIME24, AutoLogic‑cn, AutoLogic‑en \cite{zhu2025autologi}, ZebraLogic \cite{lin2025zebralogic}, LiveCodeBench \cite{jain2024livecodebench}, HumanEval\cite{chen2021evaluating}, HumanEval+ \cite{liu2023your}, and WritingBench \cite{wu2025writingbench}. For WritingBench, model performance is evaluated using the WritingBench-Critic-Model-Qwen-7B, whereas all other benchmarks are assessed with Qwen2.5-72B-Instruct. Together, these benchmarks test a wide spectrum of capabilities, including knowledge-driven reasoning, theorem-based problem solving, logical analysis, code synthesis, and open-ended writing, providing a comprehensive assessment of models’ reasoning abilities and generalization performance.

\input{tables/main-exp}
\paragraph{Answer Diversity Evaluation.}
Existing training paradigms that maximize the likelihood of ground-truth answers tend to favor reproducing exact reference expressions, which can limit the model’s propensity to generate alternative but consistent answers. To assess the diversity of model-generated answers, we analyze how probability mass is distributed across answer variants that are consistent with the reference.

Specifically, we transform the reference answers in the WebInstruct dataset into lexically distinct but consistent variants using DeepSeek‑V3 \cite{liu2024deepseek}. To ensure the reliability of these generated variants, we filter out any duplicate answers, and then manually evaluate a random sample of 2,000 rewritten answers. All sampled variants were confirmed to be valid and consistent with the original reference answers, demonstrating the effectiveness of the DeepSeek‑V3 rewriting process. At each training checkpoint, we compute the average log-likelihood assigned by the model to (i) the original ground-truth answers and (ii) the validated rewritten alternatives.

%% file: tables/main-exp.tex
\begin{table*}
    \centering
    \resizebox{\linewidth}{!}{
    \setlength\tabcolsep{2.2pt}
    \begin{tabular}{lcc|cccccc|ccc}
    \toprule
     \multirow{2}{*}{\textbf{Model}} & \multirow{2}{*}{\textbf{Base}} & \multirow{2}{*}{\textbf{Verifier}}   & \textbf{MMLU-Pro} & \textbf{GPQA} & \textbf{TheoremQA} & \textbf{MATH-500} & \textbf{Minerva} & \textbf{AIME 24} & \textbf{General} & \textbf{All}\\
 & & & Avg@2 & Avg@4 & Avg@2 & Avg@2 & Avg@2 & Avg@16 & - & -\\
 \midrule
 \multicolumn{10}{c}{\textbf{Llama Models}} \\
 \midrule
 Llama3.1-8B-Inst & Base & -- & 46.4 & 31.6 & 31.3 & 50.1 & 32.7 & 4.2 & 36.4 & 32.7 \\
 RLVR & Inst & Rule & 49.3 & 36.0 & 32.0 & 51.9 & 35.2 & 4.6 & 39.1 & 34.8  \\
RLPR & Inst & \textcolor{teal}{\ding{55}} &  \textbf{53.6} & {\ul 36.5} & {\ul 36.5} & \textbf{54.1} & \textbf{39.0} & {\ul 8.8} & {\ul 42.2} & {\ul 38.1}  \\
\rowcolor[HTML]{D7E8E8}
 DARL & Inst & \textcolor{teal}{\ding{55}} &  {\ul 52.5}  & \textbf{36.9}  & \textbf{39.4}  & {\ul 52.8}  & {\ul 37.9}  & \textbf{10.0}  & \textbf{42.9}  & \textbf{38.2}  \\
\midrule
 \multicolumn{10}{c}{\textbf{Qwen Models}} \\
 \midrule
Qwen2.5-7B & -- & -- & 45.3 & 32.4 & 41.4 & 63.0 & 37.6 & \hspace{2mm}6.5 & 39.7 & 37.7 \\
 Qwen2.5-7B-Inst & Base & -- & 54.5 & 34.2 & 47.3 & 75.4  & 49.4 & \hspace{2mm}9.4 & 45.3 & 45.0\\

\midrule
Oat-Zero & Math & Rule & 45.8 & {\ul 38.8} & 53.3 & {\ul 80.8}   & 52.1 & \textbf{29.8} & 46.0 & {\ul 50.1}\\ 	
PRIME & Math & Rule & 39.5 & 32.1 & 47.7 & 76.4  & 45.5 & 20.4 & 39.8 & 43.6 \\ 	
SimpleRL-Zoo & Math & Rule & 46.9 & 38.4 & 51.1 & 77.1  & 51.0 & {\ul 26.5} & 45.5 & 48.5  \\
\midrule
TTRL & Base & Rule & 51.1 & 34.1 & 48.8 & \textbf{82.1} & 52.8 & 15.8 & 44.7 & 47.5  \\
SimpleRL-Zoo & Base & Rule & 54.1 & 36.2 & 49.5 & 76.3 &  49.2 & 14.8 & 46.6 & 46.7 \\
RLVR & Base & Rule & 55.1 & 36.2 & 52.2 & 76.5 & 54.9 & 17.7 & 47.8 & 48.8  \\
General Reasoner & Base & Model & 55.4 & 37.4 & 52.1 & 77.0 & 51.7 & 16.0 & 48.3 & 48.3  \\
VeriFree & Base & \textcolor{teal}{\ding{55}} & 53.8 & 36.7 & 47.6 & 73.5 & 49.0 & 12.5 & 46.0 & 45.5  \\
 RLPR & Base & \textcolor{teal}{\ding{55}} &  \textbf{56.0} & 37.6 &  \textbf{55.4} & 78.0 & {\ul 56.5} & 16.3 & {\ul 49.7} & 50.0\\
\rowcolor[HTML]{D7E8E8}
 DARL & Base & \textcolor{teal}{\ding{55}} &  {\ul 55.6} & \textbf{39.4} &  {\ul 55.2} & 76.6 & \textbf{57.7} & 23.3 & \textbf{50.1} & \textbf{51.3}\\
    \bottomrule
    \end{tabular}
    } 
    \caption{Overall performance on six benchmarks. General: Average of MMLU-Pro, GPQA, TheoremQA. The highest score for each task are highlighted in {\textbf{bold}}, and the second highest are \underline{underlined}.}
    \label{tab:main_exp}
\end{table*}

%% file: chapter/7_ResultsandAnalysis.tex
\section{Results and Analysis}

\subsection{Overall Performance}

Following RLPR’s benchmark configuration, we evaluate DARL on a diverse set of reasoning and general-domain tasks. As shown in Table \ref{tab:main_exp}, our method consistently improves performance, with gains of 1.79, 1.22, and 7.03 points on GPQA, Minerva, and AIME 24, respectively, while performing comparably on MMLU-Pro and TheoremQA.

The improvements are most pronounced on tasks allowing multiple valid reasoning paths or alternative answers consistent with the reference, such as GPQA and AIME 24. By encouraging exploration of reference-consistent alternatives, DARL captures underlying reasoning structures without overfitting to a single canonical output. On tasks emphasizing factual correctness and standardized answers, such as MMLU-Pro, performance remains comparable to RLPR, showing that promoting answer diversity within reference constraints does not compromise accuracy.
\subsection{General Performance}

To evaluate the generality of our approach, we conduct experiments on diverse benchmarks spanning logical reasoning, code generation, and writing tasks, which vary in input structure, evaluation criteria, and answer openness. The results show that our method consistently improves performance across all domains, achieving average gains of 11.57 points on AutoLogic, 12.44 points on code-related benchmarks, and 3.35 points on writing tasks compared to RLPR.

Notably, the gains on writing tasks—where output diversity is inherently important for high-quality generation—demonstrate that our method effectively promotes answer diversity under reference supervision, mitigating overfitting to the provided reference answer while preserving correctness. These results further indicate that our approach improves the generalization of RLPR, yielding more robust and reliable performance across a wide range of tasks and domains without relying on domain-specific verifiers.

\input{tables/general}
\subsection{Training Dynamics of Output Diversity}
\begin{figure}[t!]
    \centering
    \includegraphics[width=1.0\linewidth]{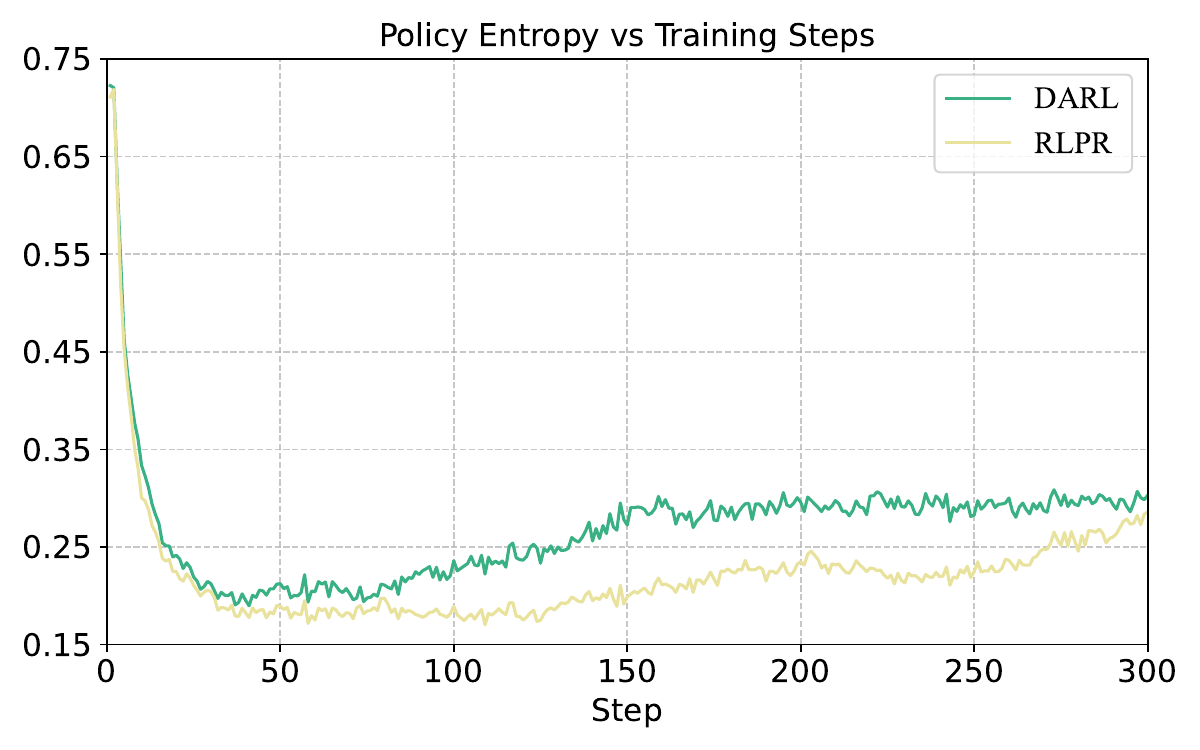}
    \caption{Policy entropy over training steps for DARL and RLPR. Our method consistently maintains higher entropy, indicating increased diversity in generated responses.}
    \label{fig:response_entropy_vs_step}
\end{figure}

To further analyze variations in model output diversity, we evaluate it from two complementary perspectives. First, we examine the evolution of policy entropy during training, which serves as an indicator of the degree of exploration and diversity in the policy over the action space. Second, we assess output-level diversity by computing the average log-likelihood assigned to a set of diverse answer candidates, which captures the model’s tendency to assign probability mass to alternative yet plausible responses.

\begin{figure}[t!]
    \centering
    \includegraphics[width=1.0\linewidth]{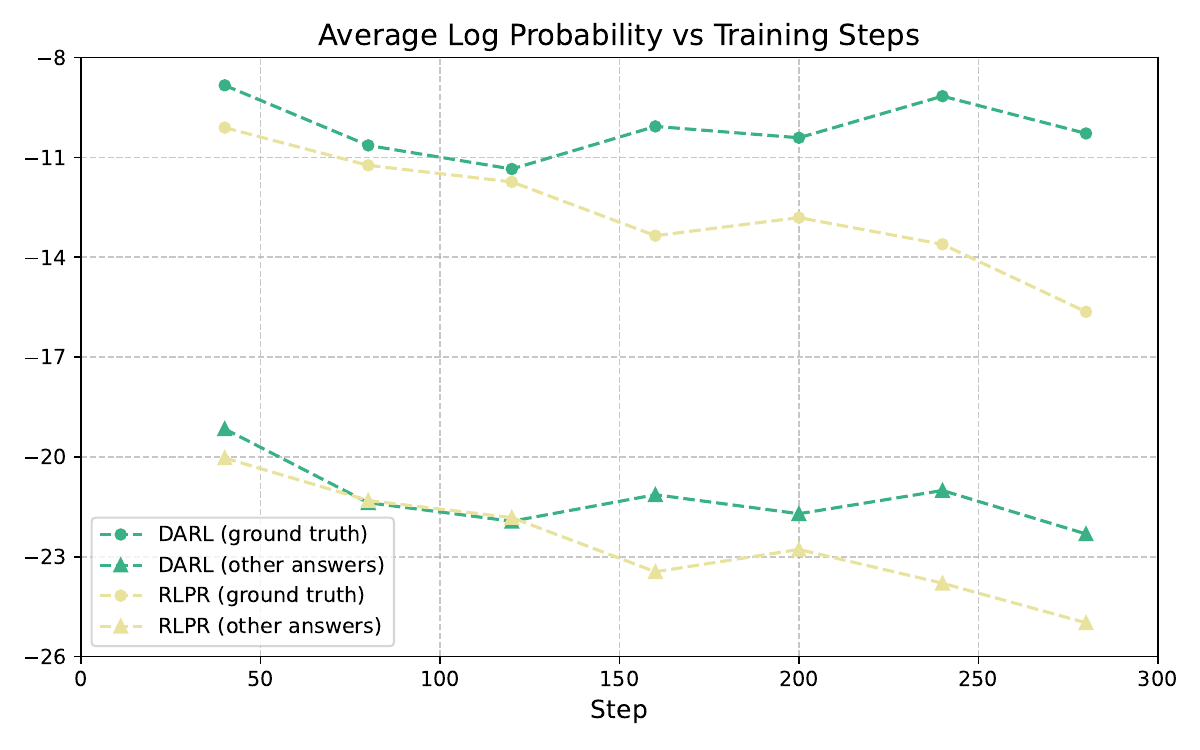}
    \caption{Average log-probability of ground-truth answers and alternative semantically equivalent answers during training. Compared to RLPR, DARL assigns higher likelihood to diverse yet semantically faithful answers.}
    \label{fig:average_log_probability_vs_step}
\end{figure}
\input{tables/verifree}
\paragraph{Our method is associated with higher policy entropy, suggesting improved exploration behavior.}
Figure~\ref{fig:response_entropy_vs_step} illustrates the evolution of policy entropy during training for both RLPR and our proposed method. In the early stage of training, the model’s capability is relatively limited, resulting in a low upper bound on the reward for exploring diverse answers. Consequently, learning is primarily driven by the RLPR reward signal, and the policy entropy of our method closely follows that of RLPR, exhibiting a rapid decline.

As training progresses into the intermediate stage, the model’s capacity gradually improves, leading to an increased upper bound on the reward associated with exploring diverse answer formulations. During this phase, our method maintains a higher policy entropy than RLPR, indicating a greater degree of exploration and output diversity.

In the later stages of training, the policy entropy of our method stabilizes, whereas that of RLPR continues to increase gradually. This divergence indicates that, although our approach encourages answer diversity, it avoids excessive randomness or output collapse. Instead, it facilitates controlled exploration within a bounded region consistent with the ground-truth answers, suggesting that the proposed diversity exploration range effectively guides the model toward diverse yet reference-consistent outputs.

\paragraph{Our method achieves higher answer diversity and yields improved performance on open-ended tasks.}

To analyze differences among the diverse answers generated by the model, we quantify output diversity by computing the average log-likelihood assigned to multiple diverse answer candidates. Specifically, for both RLPR and our proposed method, we perform rollouts of eight responses at each training step. For each response, we replace the generated answer with a set of semantically similar but lexically distinct variants and compute the corresponding log-likelihoods under the model. The average log-likelihood across these variants is then used as our answer diversity score.

The results reveal markedly different trends between RLPR and our method. As training progresses, our approach assigns higher probability mass not only to the ground-truth answers but also to alternative, semantically consistent answers, compared to RLPR. This behavior indicates that the proposed method effectively encourages the model to explore and generate diverse answer formulations while remaining aligned with the ground-truth supervision.

\subsection{Transferability to Other Verifier-Free RL Frameworks}
Beyond RLPR, our method is a general verifier-free paradigm that can be integrated into other reinforcement learning frameworks based on model-internal signals. To evaluate its transferability, we incorporate it into VeriFree and test on a diverse set of general-domain benchmarks.

The results show that adding our approach to VeriFree consistently improves performance, with an average gain of 1.2 points. This indicates that controlled exploration of consistent yet diverse answers serves as a complementary inductive bias, benefiting different verifier-free RL frameworks. By relying solely on model-internal signals and reference supervision, our method preserves the advantages of verifier-free RL while reducing overfitting to exact reference expressions, providing a practical way to enhance answer diversity and generalization across tasks.

\subsection{Ablation Study}
\input{tables/ablation1}
Table~\ref{tab:ablation} presents an ablation study on three representative benchmarks using Qwen2.5-7B. The Base Model, trained without any answer diversity reward, serves as the reference. Introducing a static answer diversity reward (SAD) consistently improves performance over the Base Model, increasing scores on AutoLogic\_en to 42.98, on HumanEval+ to 71.95, and on WritingBench to 68.16, with an overall average improvement of 1.84 points. These results indicate that incorporating answer diversity signals benefits model performance.

Replacing SAD with a dynamic answer diversity reward (DAD) further enhances results across all benchmarks, achieving the highest overall average of 62.64, which corresponds to an improvement of 3.45 points over the Base Model and 1.61 points over SAD. The gains are particularly notable on HumanEval+ with 5.49 points and on WritingBench with 3.35 points, while AutoLogic\_en also increases by 1.52 points. These findings demonstrate that dynamically adjusting the diversity reward more effectively balances the exploration of diverse answers while preserving consistency with the reference answers, yielding robust improvements across heterogeneous benchmarks. Bold numbers indicate the best-performing variant.

%% file: tables/general.tex
\begin{table*}
    \centering
    \resizebox{\linewidth}{!}{
    \setlength\tabcolsep{2.2pt}
    \begin{tabular}{lcc|ccccccc|cc}
    \toprule
     \multirow{2}{*}{\textbf{Model}} & \multirow{2}{*}{\textbf{Base}} & \multirow{2}{*}{\textbf{Verifier}}   & \multirow{2}{*}{\textbf{Autologic\_cn}} & \multirow{2}{*}{\textbf{Autologic\_en}} & \multirow{2}{*}{\textbf{Zebralogic}} & \multirow{2}{*}{\textbf{Livecodebench}} & \multirow{2}{*}{\textbf{Humaneval}} & \multirow{2}{*}{\textbf{Humaneval+}} & \multirow{2}{*}{\textbf{WritingBench}} & \multirow{2}{*}{\textbf{All}}\\
      & & &  &  &  &  &  &  &  & \\
 \midrule
 \multicolumn{10}{c}{\textbf{Llama Models}} \\
 \midrule
 Llama3.1-8B-Inst & Base & -- & 22.0  & 25.1  & \hspace{2mm}{\ul 8.9}  & {\ul 10.0}  & 61.6  & 56.7  & {\ul 66.3}  & 35.8 \\
 RLVR & Inst & Rule & \hspace{2mm}2.3  & 20.2  & 10.1  & \hspace{2mm}7.2  & 38.4  & 44.5  & 20.7  & 20.5  \\
 RLPR & Inst & \textcolor{teal}{\ding{55}} &  {\ul 27.8}  & {\ul 34.9}  & \hspace{2mm}8.3  & \hspace{2mm}9.0  & {\ul 63.4}  & \textbf{62.2}  & 53.5  & {\ul 37.0}  \\
 \rowcolor[HTML]{D7E8E8}
 DARL & Inst & \textcolor{teal}{\ding{55}} & \textbf{39.2}  & \textbf{42.0}  & \textbf{14.2}  & \textbf{11.8}  & \textbf{64.0}  & {\ul 61.6}  & \textbf{71.6}  & \textbf{43.5}  \\
\midrule
 \multicolumn{10}{c}{\textbf{Qwen Models}} \\
 \midrule
Qwen2.5-7B & -- & -- & \hspace{2mm}5.9  & \hspace{2mm}8.4  & \hspace{2mm}0.8  & \hspace{2mm}2.9  & 37.8  & 32.3  & 38.3  & 18.1 \\
 Qwen2.5-7B-Inst & Base & -- & {\ul 38.3}  & {\ul 40.6}  & \textbf{10.1}  & 12.9  & \textbf{79.3}  & {\ul 70.1}  & \textbf{80.7}  & \textbf{47.4}\\

\midrule
Oat-Zero & Math & Rule & 13.3  & 19.1  & \hspace{2mm}0.2  & \hspace{2mm}6.1  & 61.0  & 53.0  & 53.9  & 29.5\\ 	
PRIME & Math & Rule & \hspace{2mm}2.9  & \hspace{2mm}1.2  & \hspace{2mm}0.0  & \hspace{2mm}0.7  & \hspace{2mm}3.0  & \hspace{2mm}3.7  & 33.8  & \hspace{2mm}6.5 \\ 	
SimpleRL-Zoo & Math & Rule & \hspace{2mm}7.1  & 14.4  & \hspace{2mm}0.7  & \hspace{2mm}6.1  & 61.0  & 53.0  & 39.9  & 26.0  \\
\midrule
TTRL & Base & Rule & 11.1  & 12.8  & \hspace{2mm}2.1  & \hspace{2mm}3.6  & 50.6  & 44.5  & 52.0  & 25.3  \\
SimpleRL-Zoo & Base & Rule & 29.0  & 31.2  & \hspace{2mm}6.3  & \hspace{2mm}8.2  & 67.1  & 62.2  & 60.9  & 37.8 \\
RLVR & Base & Rule & 25.1  & 22.4  & \hspace{2mm}7.7  & \hspace{2mm}9.0  & 68.9  & 64.0  & 54.5  & 35.9  \\
General Reasoner & Base & Model & 29.2  & 29.0  & \hspace{2mm}8.9  & 10.0  & 61.6  & 56.7  & 66.3  & 37.4  \\
VeriFree & Base & \textcolor{teal}{\ding{55}} & 35.7  & 39.5  & \hspace{2mm}7.6  & \textbf{15.1}  & 72.0  & 65.9  & 69.5  & 43.6  \\
 RLPR & Base & \textcolor{teal}{\ding{55}} & 29.1  & 31.9  & \hspace{2mm}6.4  & \hspace{2mm}8.2  & 56.7  & 59.1  & 67.5  & 37.0\\
\rowcolor[HTML]{D7E8E8}
 DARL & Base & \textcolor{teal}{\ding{55}} & \textbf{40.3}  & \textbf{43.9}  & \hspace{2mm}{\ul 9.2}  & {\ul 13.3}  & {\ul 75.0}  & \textbf{73.2}  & {\ul 70.9}  & {\ul 46.5}\\
    \bottomrule
    \end{tabular}
    } 
    \caption{Overall performance on seven general benchmarks. The highest score for each task are highlighted in {\textbf{bold}}, and the second highest are \underline{underlined}.}
    \label{tab:general}
\end{table*}

%% file: tables/verifree.tex
\begin{table*}
    \centering
    \resizebox{\linewidth}{!}{
    \setlength\tabcolsep{2.2pt}
    \begin{tabular}{l|ccccccc|cc}
    \toprule
     {\textbf{Model}}  & {\textbf{Autologic\_cn}} & {\textbf{Autologic\_en}} & {\textbf{zebralogic}} & {\textbf{Livecodebench}} & {\textbf{Humaneval}} & {\textbf{Humaneval+}} & {\textbf{WritingBench}} & {\textbf{All}}\\
 \midrule
VeriFree & 35.7  & 39.5  & 7.6  & \textbf{15.1}  & 72.0  & 65.9  & 69.5  & 43.6  \\
\rowcolor[HTML]{D7E8E8}
 DARL & \textbf{39.3}  & \textbf{40.7}  & \textbf{9.0}  & 14.3  & \textbf{72.0}  & \textbf{68.3}  & \textbf{70.0}  & \textbf{44.8}\\
    \bottomrule
    \end{tabular}
    } 
    \caption{Overall performance on seven general reasoning benchmarks. The highest score for each task are highlighted in {\textbf{bold}}.}
    \label{tab:verifree}
\end{table*}

%% file: tables/ablation1.tex
{
\setlength{\tabcolsep}{4pt}
\begin{table}[t!]
\centering
\fontsize{7}{9}\selectfont
\begin{tabular}{l|ccc|c}
\toprule
{\centering\textbf{Method}} 
& {\textbf{Autologic}} 
& {\textbf{Humaneval+}} 
& {\textbf{WritingBench}} 
& {\textbf{Avg.}} \\
\midrule
\multicolumn{4}{c}{\textbf{Qwen2.5-7B}} \\
\midrule
$Base\ Model$
& {\cellcolorbyimprove{42.35}{42.35}\%} & {\cellcolorbyimprove{67.68}{67.68}\%} & {\cellcolorbyimprove{67.54}{67.54}\%} 
 & {\cellcolorbyimprove{59.19}{59.19}\%} \\
$w/\ SAD$
& {\cellcolorbyimprove{42.35}{42.98 }\%} & {\cellcolorbyimprove{67.68}{71.95 }\%} & {\cellcolorbyimprove{67.54}{68.16 }\%} 
 & {\cellcolorbyimprove{59.19}{61.03 }\%} \\
 $w/\ DAD$     
& {\cellcolorbyimprove{42.35}{43.87}\%} & {\cellcolorbyimprove{67.68}{73.17}\%} & {\cellcolorbyimprove{67.54}{70.89}\%} 
 & {\cellcolorbyimprove{59.19}{62.64 }\%} \\

\bottomrule
\end{tabular}
\caption{Ablation study on three representative benchmarks. Base Model is trained without any answer diversity reward. SAD denotes the static answer diversity reward, and DAD denotes the dynamic answer diversity reward. Values in \textbf{bold} indicate the best-performing variant.}
\label{tab:ablation}
\end{table}
}

%% file: chapter/8_Conclusion.tex
\section{Conclusion}
Our findings emphasize the importance of explicitly encouraging answer diversity in verifier-free reinforcement learning. By introducing a controlled diversity objective under reference supervision, DARL incentivizes models to explore multiple valid outputs rather than overfitting to a single reference answer. This diversity-aware training leads to more flexible and robust behavior, improving generalization across both reasoning and general-domain tasks. We hope this work motivates further research on incorporating explicit diversity-aware objectives into reinforcement learning for language models, particularly in open-ended and multi-step reasoning settings.

%% file: chapter/9_Limitations.tex
\section*{Limitations}
A primary limitation of DARL is its reliance on direct supervision from ground-truth answers. This dependence restricts its applicability in scenarios where only queries are available without reference answers. In such cases, rule-based verifiers or alternative methods that do not rely on reference responses may be the only viable options.

%% file: chapter/11_appendix.tex
\section{Training Details}
\input{tables/sensitivity}
\label{sec:appendix}
We train our models using the GRPO algorithm, a variant of PPO, without KL control. The training batch size is 768, with an actor mini-batch size of 192, and each step employs 8 rollouts to generate diverse reasoning trajectories. Policy updates are regulated by standard clipping coefficients (0.2 lower bound, 0.27 upper bound) to maintain training stability. Experiments are conducted on 32 GPUs with 80 GB memory each and require approximately 45 hours. Notably, our training setup closely follows that of RLPR, ensuring fair and consistent comparisons.
\input{tables/training_prompt}
\section{Parameter sensitivity analysis}

Table~\ref{tab:sensitivity} presents a sensitivity analysis of the key hyperparameters that regulate answer-diversity control in our framework, namely the diversity reward weight $\beta$ and the exploration-bound parameter $\gamma$. The objective of this analysis is to quantify how these design choices affect performance across diverse evaluation benchmarks (AutoLogic, HumanEval+, and WritingBench), and to validate the default configuration adopted in our main experiments.

The parameter $\beta$ determines the relative contribution of the answer-diversity reward in the overall optimization objective. Increasing $\beta$ amplifies the contribution of the answer-diversity reward in the optimization objective; however, excessively large values may shift the training focus toward promoting diversity rather than improving the model’s ability to generate reference answers. As shown in Table~\ref{tab:sensitivity}, enabling the diversity reward ($\beta > 0$) consistently improves the average performance over the $\beta = 0$ baseline, indicating that explicitly encouraging controlled diversity is beneficial. However, the performance gains are not monotonic with respect to $\beta$: while $\beta = 0.01$ achieves the best overall results under the default exploration bound, larger values (e.g., $\beta = 0.05$ and $\beta = 0.10$) do not yield further improvements in the average score. This trend suggests diminishing returns from stronger diversity regularization and potential overemphasis on diversity at the expense of answer quality.

The parameter $\gamma$ controls the admissible exploration region for answer diversity by specifying an effective upper bound on allowable deviation. Under our formulation, a larger $\gamma$ enforces a stricter constraint by reducing the exploration upper bound, thereby limiting the extent of diversity the model can explore. Our results indicate that $\gamma$ plays a crucial role in balancing answer diversity and reference consistency: overly restrictive constraints weaken the effect of the diversity reward, whereas excessively permissive exploration can destabilize training and degrade solution quality. Notably, a moderately relaxed constraint (e.g., $\gamma = 8$) yields higher average performance than the default setting of $\gamma = 10$, suggesting that a slightly broader exploration region allows the model to better exploit the diversity reward.

Taken together, these results indicate that model performance is sensitive to both the magnitude of the diversity reward and the scope of the exploration bounds. Careful calibration of $\beta$ and $\gamma$ is therefore crucial to achieving an effective trade-off between reference consistency and controlled diversity. This sensitivity analysis provides empirical support for our default hyperparameter choices and highlights the importance of constraining diversity exploration in verifier-free reinforcement learning.

\section{Case Study}
Figures~\ref{fig:case1-darl} and~\ref{fig:case1-rlpr} present the responses generated by DARL and RLPR, respectively, for the same question. Similarly, Figures~\ref{fig:case2-darl} and~\ref{fig:case2-rlpr} compare the responses of the two methods on another identical input.

An analysis of these responses shows that both methods produce solutions that are fully consistent in terms of logical reasoning and numerical correctness. However, DARL exhibits clear advantages over RLPR in terms of reasoning clarity and presentation quality. In particular, DARL generates responses with more standardized reasoning steps, improved readability, and a more coherent and well-structured format.
\input{tables/prompt_answer}
\input{tables/case1-balance}
\input{tables/case1-rlpr}
\input{tables/case2_balance}
\input{tables/case2_rlpr}

%% file: tables/sensitivity.tex
{
\setlength{\tabcolsep}{4pt}
\begin{table*}[t!]
\centering
\begin{tabular}{l|ccc|c}
\toprule
{\centering\textbf{Method}} 
& {\textbf{Autologic}} 
& {\textbf{humaneval+}} 
& {\textbf{WritingBench}} 
& {\textbf{Avg.}} \\
\midrule
\multicolumn{4}{c}{\textbf{DARL-7B} (Default:\ $\beta=0.01$\ and\ $\gamma=10$)} \\
\midrule
$\beta=0.00$
& {\cellcolorbyimprove{42.35}{42.35}\%} & {\cellcolorbyimprove{67.68}{67.68}\%} & {\cellcolorbyimprove{67.54}{67.54}\%} 
 & {\cellcolorbyimprove{59.19}{59.19}\%} \\
$\beta=0.01$             
& \cellcolorbyimprove{42.35}{43.87}\% & \textbf{\cellcolorbyimprove{67.68}{73.17}\%} & \textbf{\cellcolorbyimprove{67.54}{70.89}\%}
 & \textbf{\cellcolorbyimprove{59.19}{62.64 }\%} \\
\textbf{$\beta=0.05$} 
& {\cellcolorbyimprove{42.35}{43.24}\%} & {\cellcolorbyimprove{67.68}{67.68}\%} & {\cellcolorbyimprove{67.54}{69.69}\%} 
 & {\cellcolorbyimprove{59.19}{60.20}\%} \\
\textbf{$\beta=0.10$} 
 & \textbf{\cellcolorbyimprove{42.35}{46.60}\%} & \cellcolorbyimprove{67.68}{68.90}\% & \cellcolorbyimprove{67.54}{68.29}\% 
 & \cellcolorbyimprove{59.19}{61.27}\% \\
\midrule
$\gamma=12$     
& {\cellcolorbyimprove{42.35}{43.30}\%} & {\cellcolorbyimprove{67.68}{71.95}\%} & {\cellcolorbyimprove{67.54}{69.60}\%} 
 & {\cellcolorbyimprove{59.19}{61.62}\%} \\
$\gamma=10$     
& {\cellcolorbyimprove{42.35}{43.87}\%} & {\cellcolorbyimprove{67.68}{73.17}\%} & {\cellcolorbyimprove{67.54}{70.89}\%} 
 & {\cellcolorbyimprove{59.19}{62.64 }\%} \\
$\gamma=8$     
& \textbf{\cellcolorbyimprove{42.35}{45.65}\%} & \textbf{\cellcolorbyimprove{67.68}{73.17}\%} & \textbf{\cellcolorbyimprove{67.54}{71.26}\%} 
 & \textbf{\cellcolorbyimprove{59.19}{63.36}\%} \\

\bottomrule
\end{tabular}
\caption{Parameter sensitivity analysis. Cell shading reflects the relative performance improvement. \textbf{Bold} values denote the best-performing configuration.}
\label{tab:sensitivity}
\end{table*}
}

%% file: tables/training_prompt.tex
\begin{table}
\centering
\resizebox{\columnwidth}{!}{%
    \begin{tabular}{p{\columnwidth}}
    \toprule
    \textbf{Training Prompt}       \\ 
    \midrule
    A conversation between User and Assistant. The user asks a question, and the Assistant solves it. The assistant first thinks about the reasoning process in the mind and then provides the user with the answer. The reasoning process and answer are enclosed within <think> </think> and <answer> </answer> tags, respectively, i.e., <think> reasoning process here </think> <answer> answer here </answer>. \\
    \bottomrule 
    \end{tabular}
    }
\caption{Prompt used for training.}
\label{tab:trainingprompt}
\end{table}

%% file: tables/prompt_answer.tex
\begin{figure*}[htbp]
\small
\begin{tcolorbox}[title={Prompt for Rewriting Ground-Truth Answers with DeepSeek-V3}]

\textbf{Prompt:}

Rephrase the following answer into different wordings while keeping its meaning strictly the same.  
\begin{itemize}
    \item Do NOT add new information.
    \item Do NOT remove information.
    \item Keep the logical order consistent.
    \item Each version must be semantically equivalent but lexically different.
    \item Generate exactly 3 rewritten versions.
    \item Output ONLY in the following JSON format:
\end{itemize}
\[
\{
  "answer1": "...",
  "answer2": "...",
  "answer3": "..."
\}
\]

Answer to rewrite:

\end{tcolorbox}
\caption{Prompt used to rewrite ground-truth answers into semantically equivalent but lexically distinct variants with DeepSeek-V3.}
\label{fig:case1-rlpr}
\end{figure*}

%% file: tables/case1-balance.tex
\begin{figure*}[htbp]
\small
\begin{tcolorbox}[title={Case Study of DARL on WebInstruct Dataset}]

\textbf{Question:}

A quantum system (not a hydrogen atom) has energy levels
\[
E_{n=1}=1\,\text{eV},\quad
E_{n=2}=4\,\text{eV},\quad
E_{n=3}=9\,\text{eV},\quad
E_{n=4}=16\,\text{eV}.
\]
The system is initially in its lowest energy state. After absorbing a photon, it is in the \(n=3\) state.
What is the wavelength of the absorbed photon?

\rule[3pt]{\textwidth}{0.20mm}

\textbf{Reference:}

154.55 nm

\rule[3pt]{\textwidth}{0.20mm}

\textbf{DARL Response:}

<think>

To determine the wavelength of the photon absorbed by the quantum system, we first find the energy difference between the initial and final states.

The energy levels are:
\[
E_{n=1}=1\,\text{eV},\quad
E_{n=2}=4\,\text{eV},\quad
E_{n=3}=9\,\text{eV},\quad
E_{n=4}=16\,\text{eV}.
\]

The system starts in the lowest energy state \(n=1\) and ends in the \(n=3\) state, so the absorbed photon energy is
\[
\Delta E = E_{3}-E_{1}=9\,\text{eV}-1\,\text{eV}=8\,\text{eV}.
\]

Using the photon energy relation,
\[
E=\frac{hc}{\lambda}\quad\Rightarrow\quad \lambda=\frac{hc}{E}.
\]

Convert \(8\,\text{eV}\) to joules using \(1\,\text{eV}=1.6\times 10^{-19}\,\text{J}\):
\[
E = 8\times 1.6\times 10^{-19}\,\text{J} = 1.28\times 10^{-18}\,\text{J}.
\]

Now substitute \(h=6.626\times 10^{-34}\,\text{J}\cdot\text{s}\) and \(c=3.0\times 10^{8}\,\text{m/s}\):
\[
\lambda=\frac{(6.626\times 10^{-34})(3.0\times 10^{8})}{1.28\times 10^{-18}}.
\]

Compute the numerator:
\[
hc=(6.626\times 10^{-34})(3.0\times 10^{8})
=1.9878\times 10^{-25}\,\text{J}\cdot\text{m}.
\]

Thus,
\[
\lambda=\frac{1.9878\times 10^{-25}}{1.28\times 10^{-18}}
=1.55\times 10^{-7}\,\text{m}.
\]

Therefore, the wavelength of the absorbed photon is approximately \(1.55\times 10^{-7}\,\text{m}\). </think>

<answer> 1.55e-7 m </answer>

\end{tcolorbox}
\caption{Case Study of DARL on the WebInstruct Dataset.}
\label{fig:case1-darl}
\end{figure*}

%% file: tables/case1-rlpr.tex
\begin{figure*}[htbp]
\small
\begin{tcolorbox}[size=small,title={Case Study of RLPR on WebInstruct Dataset}]

\textbf{Question:}

A quantum system (not a hydrogen atom) has energy levels
\[
E_{n=1}=1\,\text{eV},\quad
E_{n=2}=4\,\text{eV},\quad
E_{n=3}=9\,\text{eV},\quad
E_{n=4}=16\,\text{eV}.
\]
The system is initially in its lowest energy state. After absorbing a photon, it is in the \(n=3\) state.
What is the wavelength of the absorbed photon?

\rule[3pt]{\textwidth}{0.20mm}

\textbf{Reference:}

154.55 nm

\rule[3pt]{\textwidth}{0.20mm}

\textbf{RLPR Response:}

<think>

To find the wavelength of the photon absorbed by the quantum system, we need to determine the energy difference between the initial and final states and then use the relationship between the photon's energy and its wavelength.

1. \textbf{Identify the initial and final states:}
   \begin{itemize}[topsep=2pt, partopsep=2pt]
     \item The system starts in the lowest energy state, which is \( n = 1 \).
     \item After absorbing a photon, the system is in the \( n = 3 \) state.
   \end{itemize}

2. \textbf{Calculate the energy difference:}
   \begin{itemize}[topsep=2pt, partopsep=2pt]
     \item The energy level for the initial state \( n = 1 \) is \( E_1 = 1 \, \text{eV} \).
     \item The energy level for the final state \( n = 3 \) is \( E_3 = 9 \, \text{eV} \).
     \item The energy difference \( \Delta E \) is:
     \[
     \Delta E = E_3 - E_1 = 9 \, \text{eV} - 1 \, \text{eV} = 8 \, \text{eV}
     \]
     So, the photon absorbed must have an energy of \( 8 \, \text{eV} \).
   \end{itemize}

3. \textbf{Use the relationship between photon energy and wavelength:}
   \begin{itemize}[topsep=2pt, partopsep=2pt]
     \item The energy of a photon is given by the equation:
     \[
     E = \frac{hc}{\lambda}
     \]
     where \( h \) is Planck's constant \( 6.626 \times 10^{-34} \, \text{J} \cdot \text{s} \), \( c \) is the speed of light \( 3.00 \times 10^8 \, \text{m/s} \), and \( \lambda \) is the wavelength of the photon.
     \item We know the energy of the photon is \( 8 \, \text{eV} \). First, convert this energy into joules. Since \( 1 \, \text{eV} = 1.6 \times 10^{-19} \, \text{J} \), we have:
     \[
     8 \, \text{eV} = 8 \times 1.6 \times 10^{-19} \, \text{J} = 1.28 \times 10^{-18} \, \text{J}
     \]
     \item Using the equation for the energy of a photon \( E = \frac{hc}{\lambda} \), we can solve for the wavelength \( \lambda \):
     \[
     \lambda = \frac{hc}{E}
     \]
     \item Substitute in the values for \( h \), \( c \), and \( E \):
     \[
     \lambda = \frac{(6.626 \times 10^{-34} \, \text{J} \cdot \text{s})(3.00 \times 10^8 \, \text{m/s})}{1.28 \times 10^{-18} \, \text{J}}
     \]
     \item First, calculate the numerator:
     \[
     (6.626 \times 10^{-34})(3.00 \times 10^8) = 1.987 \times 10^{-25} \, \text{J} \cdot \text{m}
     \]
     \item Now, divide the numerator by the photon energy:
     \[
     \lambda = \frac{1.987 \times 10^{-25} \, \text{J} \cdot \text{m}}{1.28 \times 10^{-18} \, \text{J}} = 1.55 \times 10^{-7} \, \text{m}
     \]
   \end{itemize}
So, the wavelength of the photon absorbed will be approximately \( 1.55 \times 10^{-7} \, \text{m} \) or \( 155 \, \text{nm} \).</think>

<answer>\( 155 \, \text{nm} \)</answer>

\end{tcolorbox}
\caption{Case Study of RLPR on the WebInstruct Dataset.}
\label{fig:case1-rlpr}
\end{figure*}

%% file: tables/case2_balance.tex
\begin{figure*}[htbp]
\small
\begin{tcolorbox}[title={Case Study of DARL on WebInstruct Dataset}]

\textbf{Question:}

The equilibrium constant, \( K_c \), for the following reaction is \( 7.00 \times 10^{-5} \) at 673 K.

\[
NH_4I(s) \to NH_3(g) + HI(g)
\]

Calculate the equilibrium concentration of \( HI \) when 0.590 moles of \( NH_4I(s) \) are introduced into a 1.00 L vessel at 673 K.

\rule[3pt]{\textwidth}{0.20mm}

\textbf{Reference:}

0.00837 M

\rule[3pt]{\textwidth}{0.20mm}

\textbf{DARL Response:}

<think>

To determine the equilibrium concentration of HI when 0.590 moles of \( \text{NH}_4\text{I}(s) \) are introduced into a 1.00 L vessel at 673 K, we'll follow these steps:

1. \textbf{Write the balanced equation and initial conditions:}
   The reaction is:
   \[
   \text{NH}_4\text{I}(s) \rightleftharpoons \text{NH}_3(g) + \text{HI}(g)
   \]
   Since \( \text{NH}_4\text{I}(s) \) is a solid, its concentration does not appear in the equilibrium expression. The equilibrium expression is:
   \[
   K_c = \frac{[\text{NH}_3][\text{HI}]}{1} = [\text{NH}_3][\text{HI}]
   \]
   The value of \( K_c \) is given as \( 7.00 \times 10^{-5} \).

2. \textbf{Set up the initial concentrations:}
   Initially, there are 0.590 moles of \( \text{NH}_4\text{I}(s) \) in a 1.00 L vessel, which means initially there is no \( \text{NH}_3(g) \) and no \( \text{HI}(g) \), so the initial concentrations of \( \text{NH}_3 \) and \( \text{HI} \) are both 0 M.

   When \( \text{NH}_4\text{I} \) starts to decompose, let \( x \) be the amount of \( \text{NH}_3 \) and \( \text{HI} \) formed at equilibrium. Then:
   \[
   \text{NH}_4\text{I}(s) \rightleftharpoons \text{NH}_3(g) + \text{HI}(g)
   \]
   The changes in concentrations will be:
   \[
   \begin{array}{|c|c|c|c|}
   \hline
   & \text{NH}_4\text{I}(s) & \text{NH}_3(g) & \text{HI}(g) \\
   \hline
   \text{Initial (M)} & - & 0 & 0 \\
   \text{Change (M)} & - & +x & +x \\
   \text{Equilibrium (M)} & - & x & x \\
   \hline
   \end{array}
   \]
   The equilibrium concentrations are thus:
   \[
   [\text{NH}_3] = x \quad \text{and} \quad [\text{HI}] = x
   \]

3. \textbf{Substitute into the equilibrium expression:}
   Using the expression for \( K_c \):
   \[
   K_c = [\text{NH}_3][\text{HI}] = x \cdot x = x^2
   \]
   Given that \( K_c = 7.00 \times 10^{-5} \), we have:
   \[
   x^2 = 7.00 \times 10^{-5}
   \]

4. \textbf{Solve for \( x \):}
   Taking the square root of both sides:
   \[
   x = \sqrt{7.00 \times 10^{-5}}
   \]
   Calculating the square root:
   \[
   x = \sqrt{7.00 \times 10^{-5}} = 8.37 \times 10^{-3}
   \]

5. \textbf{Determine the equilibrium concentration of HI:}
   The equilibrium concentration of HI is equal to \( x \), which is \( 8.37 \times 10^{-3} \) M.

Therefore, the equilibrium concentration of HI is approximately \( 8.37 \times 10^{-3} \) M.</think>

<answer> \( 8.37 \times 10^{-3} \) M </answer>

\end{tcolorbox}
\caption{Case Study of DARL on the WebInstruct Dataset.}
\label{fig:case2-darl}
\end{figure*}

%% file: tables/case2_rlpr.tex
\begin{figure*}[htbp]
\small
\begin{tcolorbox}[title={Case Study of RLPR on WebInstruct Dataset}]

\textbf{Question:}

The equilibrium constant, \( K_c \), for the following reaction is \( 7.00 \times 10^{-5} \) at 673 K.

\[
NH_4I(s) \to NH_3(g) + HI(g)
\]

Calculate the equilibrium concentration of \( HI \) when 0.590 moles of \( NH_4I(s) \) are introduced into a 1.00 L vessel at 673 K.

\rule[3pt]{\textwidth}{0.20mm}

\textbf{Reference:}

0.00837 M

\rule[3pt]{\textwidth}{0.20mm}

\textbf{RLPR Response:}

<think>

To find the equilibrium concentration of HI when 0.590 moles of NH4I(s) are introduced into a 1.00 L vessel at 673 K, and we know the equilibrium constant \( K_c = 7.00 \times 10^{-5} \) for the reaction:

\[
\text{NH}_4I(s) \rightarrow \text{NH}_3(g) + \text{HI}(g)
\]

First, note that the reaction involves the dissociation of solid NH4I into gaseous NH3 and HI. The concentration of solids does not appear in the equilibrium expression because their concentration is essentially constant and does not change during the reaction.

The balanced chemical reaction is:

\[
\text{NH}_4I(s) \rightarrow \text{NH}_3(g) + \text{HI}(g)
\]

Since NH4I is a solid, we only need to focus on the concentrations of NH3 and HI at equilibrium. Let's denote the change in the amount of NH3 and HI at equilibrium as \( x \) moles of NH3 and HI formed at equilibrium.

We start with 0.590 moles of NH4I(s). Since it's a solid and its amount will not change directly in the equilibrium equation, we can focus on the gaseous products.

At equilibrium, the amount of NH3 and HI will be \( x \) moles each.

The initial amount of NH3 and HI at the start of the reaction is 0 moles because they are formed from the dissociation of NH4I.

At equilibrium, the amount of NH3 and HI will both be x moles. Since the reaction starts with 0.590 moles of NH4I(s) and the reaction is 1:1:1, the moles of NH3 and HI at equilibrium will be x moles each.

Since the volume of the container is 1.00 L, the equilibrium concentrations will be:

\[
[\text{NH}_3] = x \, \text{M}
\]
\[
[\text{HI}] = x \, \text{M}
\]

The equilibrium expression for the reaction will be:

\[
K_c = [\text{NH}_3][\text{HI}]
\]

We know from the problem that:

\[
K_c = 7.00 \times 10^{-5}
\]

Substituting the equilibrium concentrations into the \( K_c \) expression:

\[
7.00 \times 10^{-5} = x \cdot x = x^2
\]

To find \( x \), we take the square root of both sides:

\[
x = \sqrt{7.00 \times 10^{-5}}
\]

Calculating the square root:

\[
x = \sqrt{7.00 \times 10^{-5}} \approx 0.00837 \, \text{M}
\]

Thus, the equilibrium concentration of HI will be \( x \).

So, the equilibrium concentration of HI will be \( 0.00837 \, \text{M} \).

Therefore, the answer is: 0.00837 M</think>

<answer>0.00837 M</answer>

\end{tcolorbox}
\caption{Case Study of RLPR on the WebInstruct Dataset.}
\label{fig:case2-rlpr}
\end{figure*}